\documentclass[runningheads]{llncs}
\usepackage{graphicx}

\usepackage{microtype}
\usepackage{subfigure}
\usepackage{float}
\usepackage{booktabs} 

\usepackage{amssymb}
\usepackage{hyperref}
\PassOptionsToPackage{hyphens}{url}
\usepackage{lipsum}
\usepackage{physics}
\usepackage{tikz}
\usetikzlibrary{automata,positioning}

\newcommand{\be}{\begin{eqnarray}}
\newcommand{\ee}{\end{eqnarray}}

\usepackage{xcolor}
\usepackage{tikz,xcolor,hyperref}

\definecolor{lime}{HTML}{A6CE39}
\DeclareRobustCommand{\orcidicon}{
	\begin{tikzpicture}
	\draw[lime, fill=lime] (0,0) 
	circle [radius=0.16] 
	node[white] {{\fontfamily{qag}\selectfont \tiny ID}};
	\draw[white, fill=white] (-0.0625,0.095) 
	circle [radius=0.007];
	\end{tikzpicture}
	\hspace{-2mm}
}

\foreach \x in {A, ..., Z}{%
	\expandafter\xdef\csname orcid\x\endcsname{\noexpand\href{https://orcid.org/\csname orcidauthor\x\endcsname}{\noexpand\orcidicon}}
}

\begin{document}

\title{Predicting Traffic Accident Severity\\with Deep Neural Networks}
\titlerunning{Predicting Traffic Accident Severity with DNNs}

\author{Meghan Bibb
\inst{1}
\and
Pablo Rivas
\inst{2}\orcidB{} 
\and
Mahee Tayba
\inst{3}
}

\institute{
School of Engineering and Computer Science \\
Department of Computer Science \\
Baylor University, Texas, USA\\
$^1$\email{Meghan\_Bibb1@baylor.edu}
$^2$\email{Pablo\_Rivas@baylor.edu}
$^3$\email{Mahee\_Tayba1@baylor.edu}
}
\maketitle
\begin{abstract}
Traffic accidents can be studied to mitigate the risk of further events. Recent advances in machine learning have provided an alternative way to study data associated with traffic accidents. New models achieve good generalization and high predictive power over imbalanced data. In this research, we study neural network-based models on data related to traffic accidents. We begin analyzing relative feature colinearity and unsupervised dimensionality reduction through autoencoders, followed by a dense network. The features are related to traffic accident data and the target is to classify accident severity. Our experiments show cross-validated results of up to 92\% accuracy when classifying accident severity using the proposed deep neural network.
\end{abstract}

\section{Introduction}

Traffic accidents are tragic causes of death and will be the seventh largest cause of death by 2030, according to recent studies \cite{RoadTrafficAccidentSeveritySouthAfrica,chen2019global}. Investigating the nature of traffic accidents and what causes their severity is crucial to building better and safer transportation infrastructure and legislation \cite{sheng2018effect,castillo2019legislation,favretto2018driving}. Much research has been done in analyzing and predicting the severity of accidents \cite{ihueze2018road}, including deep learning approaches \cite{ren2018deep,sameen2017severity}, or fuzzy logic models \cite{gaber2017traffic}. Of particular interest was the research done at The Ohio State University to predict accident risk \cite{CornellCitation2}. Their research predictions of accident severity is done with data that was gathered after the accident had occurred, such as the description of the accident or the duration length of the accident, which makes it a unique approach. This paper builds on a similar idea, but we approach the problem from a different perspective. We aim to use a machine learning model to predict the severity of an accident provided the conditions \textit{before} the accident occurred. This can be accomplished by removing features from consideration that are only observed during or after the accident has happened. Similar research has focused in building predictions based on large scale data sets of traffic data \cite{BigDataTraffic,CornellCitation1}. However, as with many big data problems, high dimensionality is a frequent problem in such models, some considering over 100 raw features. This paper also uses a technique for dimensionality reduction to make traffic accident severity predictions, mitigating the curse dimensionality for simpler machine learning models such as the proposed deep autoencoder model. Preliminary results suggest that using an autoencoder, for dimensionality reduction yields promising results of up to 92\% accuracy.

This paper is organized as follows: Section II describes the dataset used in this research. Section III presents a brief review of the state of the art germane to the problem of accident risk prediction and unsupervised machine learning on traffic data. The methodology is addressed in Section IV while Section V describes the experimental setup and models. Section VI discusses the results while Section VII draws important conclusions for this paper.

\section{Dataset Overview}
\begin{figure*}[!b]
\centering
\begin{minipage}[t]{0.49\textwidth}
  \includegraphics[width=\linewidth]{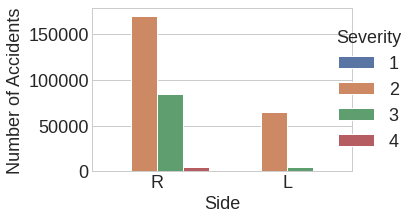}
  \caption{Distribution of side [of street] \newline per severity level.}
  \label{fig:Side}
\end{minipage} \hfill
\begin{minipage}[t]{0.49\textwidth}
  \includegraphics[width=\linewidth]{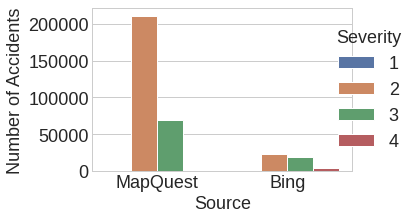}
  \caption{Distribution of source per severity level.}
  \label{fig:Source}
\end{minipage}
\end{figure*}

\begin{figure*}[!t]
\begin{minipage}[t]{0.49\textwidth}
  \includegraphics[width=\linewidth]{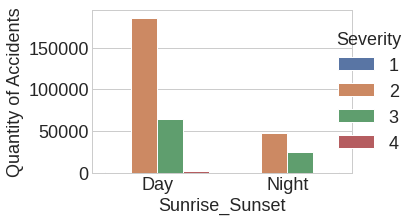}
  \caption{Distribution of daytime and \newline nighttime accidents.}
 \label{fig:Sunrise_Sunset}
\end{minipage} \hfill
\begin{minipage}[t]{0.49\textwidth}
  \includegraphics[width=\linewidth]{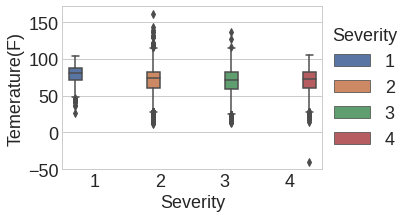}
  \caption{Distribution of temperature (F) per severity level.}
 \label{fig:Temp}
\end{minipage}
\end{figure*}

\begin{figure*}[!t]
\centering
\begin{minipage}[t]{0.49\textwidth}
  \includegraphics[width=\linewidth]{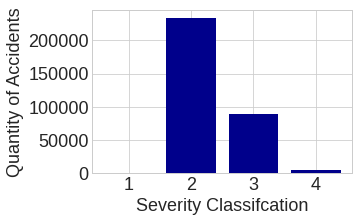}
  \caption{Distribution of accident severity.}
  \label{fig:Severity}
\end{minipage} \hfill
\begin{minipage}[t]{0.49\textwidth}
  \includegraphics[width=\linewidth]{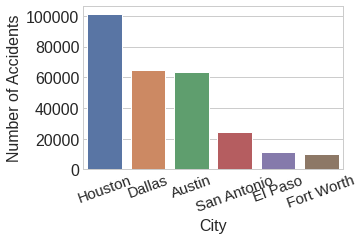}
  \caption{Top six cities represented.}
  \label{fig:top6}
\end{minipage}
\end{figure*}

\begin{figure}[th!]
  \centering
  \includegraphics[width=0.45\columnwidth]{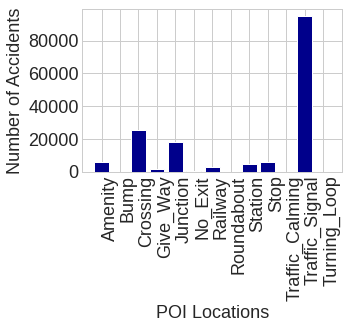}
  \caption{Distribution of points of interest. \centering}
  \label{fig:POI}
\end{figure}

The data used in this project was collected from MapQuest and Microsoft Bing. It contains 3.5 million records of accidents across the contiguous United States beginning in 2016 and ending in the summer of 2020 \cite{CornellCitation1,CornellCitation2}. This paper focuses on accidents in Texas, reducing the size of the data to 330,000 accidents. The data set has 49 features including severity which ranges from one to four, with four being the most severe and one being the least. The data set is made up of 0.3\% severity level one instances, 71.0\% severity level two instances, 27.2\% severity level three instances, and 1.4\% severity level four instances. Additionally this data set contains traffic attributes, address attributes, weather attributes, points of interest attributes, and information regarding the period of day. Fig. \ref{fig:Severity} and Fig. \ref{fig:POI} show the distributions of some of the features of the data set. Fig. \ref{fig:Side},  Fig. \ref{fig:Source}, Fig. \ref{fig:Sunrise_Sunset}, and Fig. \ref{fig:Temp} show the distributions of features after accident records have been split by severity. The full data set can be viewed at  {\tt https://smoosavi.org/datasets/us\_accidents} \cite{CornellCitation1,CornellCitation2}.

\section{Background and Related Work}
\subsection{Accident Risk Prediction}

Naturally, due to the tragic and universal nature of traffic accidents a lot of work has gone into studying and predicting traffic accidents. Research has been done to predict traffic severity using dash-cam footage \cite{10.1145/3356995.3364535}, real-time accident prediction \cite{8842686,5203263}, accident prediction in terms of POI structure \cite{5620615}, a variety of neural networks for predictive models \cite{8056908,CornellCitation2,10.1145/3357384.3357829,8687542}, and much more research.

The work in this paper is largely built off of the work completed in Accident Risk Prediction based on Heterogeneous Sparse Data \cite{CornellCitation2}. These researchers built models for traffic severity in six cities using a feature vector containing data in five categories, traffic, time, weather, points of interest (POI), and natural language. The researchers were very effective in detecting severity of accidents in these six cities by implementing a deep-neural-network-based accident prediction (DAP) model  \cite{CornellCitation2}. This DAP model used 24 time-variant attributes and 113 time-invariant attributes. This paper's approach differs in three key ways from this paper \cite{CornellCitation2}. Only data that could be gathered before an accident occurs (time-invariant) will be used in this project. This will cause natural language and time-variant data to not be used. Greater dimensionality reduction will also be incorporated into this project to allow the models built to rely on less features while striving for similar success metrics. Additionally, data from all cities and locations in Texas will be included in the data set.

\subsection{Autoencoders and Traffic Data}

Recent research has also utilized autoencoders for use with traffic data. Specifically, researchers used autoencoders with  traffic data from Kerala to perform feature selection \cite{9225336}, and to represent traffic data using encodings \cite{8687542}. Additionally, research has been done to predict accidents by finding reconstruction error of embedded video footage \cite{8367975}. Stack Denoising Convolutional Auto-Encoder (SDCAE) was developed to determine accident risk within Chinese cities \cite{8530861}. It shows that the use of autoencoders improved upon traditional machine learning methods due to the complex nature of traffic data. However, this paper differs from \cite{8530861} as our paper predicts the severity of an accident whereas in \cite{8530861} the authors predicts whether or not an accident will occur within a city. 

\section{Methodology}

After building data visualizations to gain a better understanding of the data’s distributions, we wanted to find the correlation between each of the 49 features and severity. Due to the data set having a large quantity of categorical variables a measure of association other than Pearson's correlation was needed. However, using Cramer’s V, we were able to measure the association between categorical features to other categorical features as well as to numerical features. Cramer’s V has a range between 0 and 1 with values of 0.1 indicating a moderate association and values above 0.15 indicating a strong connection \cite{CramersV}. Association matrices to visualize these relationships were created using the Dython python library.

As the data set presents 49 features, many of these features may not have a relationship with the level of severity of an accident. To find the features that have the highest correlation with severity levels, this requires comparing severity levels to each feature. Performing dimensionality reduction can eliminate features unrelated to severity and reduce the curse of dimensionality. 

After we worked to reduce the quantity of features being considered, the data needed to prepared to work with the models in the experiments. We used sklearn's StandardScaler on all numeric values to normalize the data and create a mean of zero. Additionally, the categorical variables were one-hot encoded to transform them into numeric binary values.
While initial dimensionality reduction reduced the amount of features in the data set, after one-hot encoding many of the remaining categorical values the dimensions of the data became quite large. To further work to reduce dimensionality an autoencoder was created to encode the remaining features into a smaller latent dimension \cite{bahadur2019dimension,rivas2019deep}. This reduced the dimensions of the data set from 1218 to 256. 

Finally, two main experiments were built. A deep neural network was created with the reduced features from the correlation experiments to predict the level of severity of a potential accident. All experiments were written in Python using Tensorflow,  scikit-learn, and Keras. Additionally, a deep neural network was created with the input of reduced features transformed by the encoder to a dimension of 256. Tensorboard was then used to tune the hyperparameters of each model, creating 54 models for comparison for each experiment. By finding the best hyperparameters for each model we then were able to compare the two models after retraining each with the optimized hyperparameters.

\section{Experiments}

This section explains our experiments in working with dimensionality reduction, building an autoencoder, and building DNNs to predict accident severity.

\subsection{Dimensionality Reduction}
By calculating Cramer’s V for each feature’s association with severity we learned which features would best help in the prediction of accident severity \cite{CramersV}. The experiment showed the location of the accident is most associated with the severity followed by if the accident took place at a traffic signal, and the side of the street the car was on.

\subsection{The Autoencoder}
The autoencoder was trained using the features identified at or above a 0.2 Cramer’s V association threshold. Geographic categorical features of county, city, and airport code as well as numeric features of starting Latitude and starting Longitude were used as input to the autoencoder. All points of interest were included and the side of street the accident took place on were also included as input. Each categorical feature was one-hot encoded and each numeric feature was scaled using sklearn's StandardScaler. Using 60\% of the data for training, 20\% for testing, and 20\% for validation, the deep autoencoder was trained for 200 epochs with a batch size of 1000.

\begin{figure}[!t]
\centering
  \includegraphics[width=0.7\columnwidth]{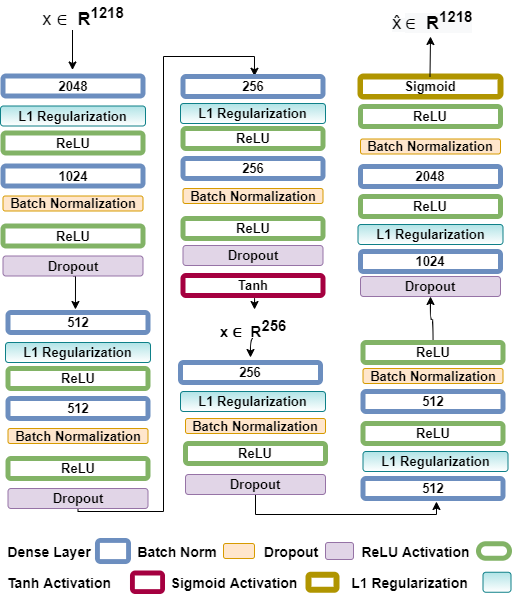}
  \caption{The architecture for the autoencoder}
  \label{fig:AEArch}
\end{figure}

\subsection{The Deep Neural Network}

Two deep neural networks were built to run this project's main tests on. The first took the same input as the autoencoder. The second was trained with the encoded training data from the autoencoder's latent space encoder model's predictions. This caused the second model to use input with roughly one fourth of the dimensionality of the first model.
The training, test, and validation data remained the same with the same 60:20:20 split, using sklearn to ensure that each had the same ratio of severity levels in each data set.
Because of the unequal distribution of severity levels, The experiments focused on accurately predicting the lesser represented severity levels one and four just as well as the heavily represented levels two and three. To do this, balanced class weights were created based on the distribution of severity to be used in training the model. The class weights consisted of 75.94 for a severity level of one, 0.35 for a severity level of two, 0.92 for a severity level of three, and 17.49 for a severity level of four.

To find the best hyperparameters for each deep neural network, Tensorboard was used for hyper-parameter tuning, utilizing the HParam dashboard. The number of initial neurons (1218, 2436, 3654), the initial dropout rate (0.2, 0.3, 0.4), the batch size (2000, 5000, 10000), and the l2 regularizer penalty factor (0.001, 0.0001) were compared. 
These comparisons generated every combination of the values provided generating 54 total models. BER was used as a comparison to calculate the average error rate due to severity's imbalance. 

\begin{figure}[!t]
\centering
  \includegraphics[width=0.7\linewidth]{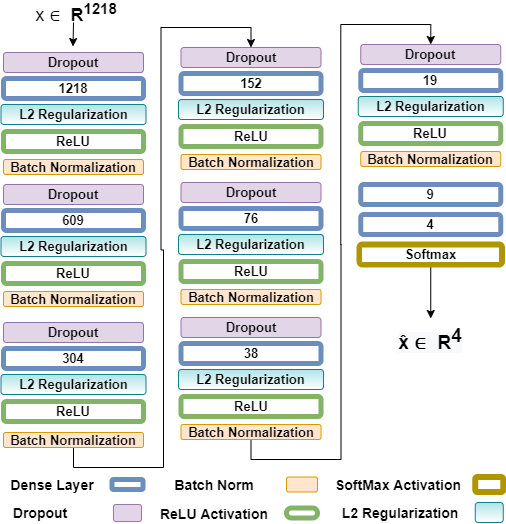}
  \caption{The architecture for the deep neural network}
  \label{fig:DeepNeuralNetArchitecture}
\end{figure}

The hyper-parameter tests with the model without encoder input showed that 1218 initial neurons, a dropout rate of 0.3, a batch size of 5000 and a l2 regularizer penalty of 0.0001 generated the model with the best metrics. Fig. \ref{fig:DeepNeuralNetArchitecture} displays the architecture of this model. Cross validation was then used to repeat the experiment 10 times with these optimized hyperparameters, with an average BER of 0.3575 with a standard deviation of 0.069. 
The hyper-parameter tests with the model with the encoder input showed that 1218 initial neurons, a dropout rate of 0.2, a batch size of 5000 and a l2 regularizer penalty of 0.0001 generated the best metrics. Fig. \ref{fig:AEArch} displays the architecture of this model. Cross validation was then used to repeat the experiment 10 times with these optimized hyperparameters, with an average BER of 0.4095 with a standard deviation of 0.067.
Finally, we wanted to ensure that the use class weights had actually been effective in decreasing the BER for detecting severity. By using an identical model to the one generated out of hyperparameter testing with no encoder input we removed the use of class weights. With all other parameters remaining a constant we performed a 10-fold cross-validation. The BER for the model without class weights rose to 0.6662 with a standard  deviation  of 0.022. 

\section{Discussion and Analysis}

We will now discuss the results of the autoencoder and DNN experiments.

\subsection{The Autoencoder}

\begin{figure}[!t]
\centering
  \includegraphics[width=0.8\columnwidth]{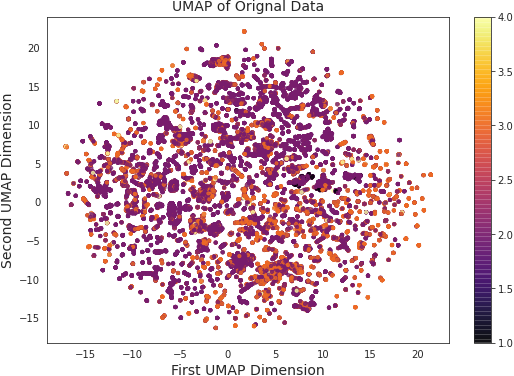}
  \caption{The original validation data}
  \label{fig:originalNoAE}
\end{figure}

\begin{figure}[!t]
\centering
  \includegraphics[width=0.8\columnwidth]{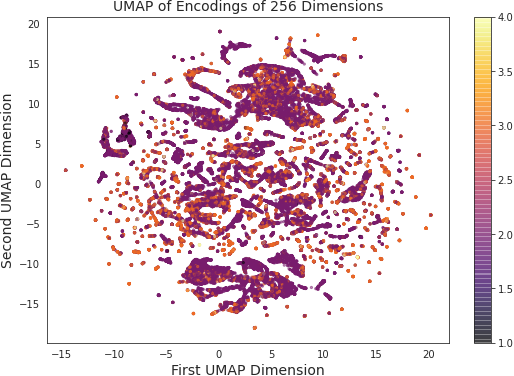}
  \caption{The encoded validation data}
  \label{fig:validationAfterAE}
\end{figure}

Fig. \ref{fig:originalNoAE} displays the results of the original validation data mapped into two dimensions by UMAPS \cite{mcinnes2020umap}. While we can see some clusters of accidents with a severity of 2, most of the data is well mixed, especially the majority of the data in the middle. Fig. \ref{fig:validationAfterAE} contains the same validation data after it has been transformed by the encoder portion of the trained autoencoder. While there were no explicit separation of severity levels, the autoencoder shows a less uniform distribution, spreading the points out into more clearly defined clusters. This is evident when observing the centers of each UMAP display, where the encoded UMAP plot shows more white space and shape in the center whereas the original plot is very condensed and not separable. By being able to show the same, if not more, patterns and shapes in the data with roughly one fourth of the dimensions, the autoencoder is successful enough to be used in the DNN experiments.

\subsection{The Deep Neural Network}

The deep neural network with encoded input had a BER greater than the model using unmodified input by 0.0520. This result was surprising to us, as we thought that with the autoencoder encoding the data into less dimensions while still being able to represent the data, it would decrease the BER more than the deep neural network with the unmodified data.  
While both models maintained low BER rates, we wanted to see the effect of the use of class weights on the BER from the cross-validation experiments. We ran the same cross-validated experiment with the best performing of the two models, the model without the encoder. As the model with no initially assigned class weights ran, it showed that the model's BER rose to 0.6662 with a standard deviation of .022. Additionally, not a single one of the model's 329,284 predictions over the validation data set was predicted as a severity level of one. This great increase in BER showed that the class weights were critical to identifying instances of imbalanced classes in this model. The ability to classify imbalanced classes, especially the smallest represented severity class of one, comes at the cost of accuracy. Accuracy rose from 66.7\% to 76.2\% when initial class weights were removed. This is due to the model's ability to make better predictions on heavily represented severity levels such as two, whose accuracy in predictions rose from 65.6\% to 92.1\% with the removal of initial class weights. A summary of the performance results is shown in Table \ref{tab:results}.

\begin{table}[t!]
    \centering
    \caption{Performance comparison between models}
    \begin{tabular}{l|rl|rl}
    \hline\hline
        Model & BER & $\sigma$ & ACC & $\sigma$ \\ \hline
        Encoder + DNN & 0.4095 & 0.067 & 0.6216 & 0.017 \\
        DNN & 0.3575 & 0.069 & 0.6665 & 0.097  \\
    \hline\hline
    \end{tabular}
    \label{tab:results}
\end{table}

\subsection{Applications}
Traffic accidents are a substantial public safety concern because of the high death toll and economic losses caused by them each year around the world. Our proposed deep neural network model is a promising traffic accident prediction tool with various practical applications. This model uses the underlying correlations between contextual information such as location, weather condition, type of intersection while predicting future traffic accidents. Therefore it can be helpful to landscape architects and civil engineers to design highways and streets. Combined with big data, this method can be applied to improve traditional public transport management systems by building a safety risk assessment system and smart traffic location information system. There is also a scope to include imagery information to increase our model's predictive capacity and anticipate where traffic accidents are likely to occur.

\section{Conclusion}
This paper proposes a deep neural network model that predicts accident severity based on features that could be observed before an accident. By using correlation and association measures the most relevant features are chosen for use. Then, a deep autoencoder was used to create an encoder for dimensionality reduction to gather compressed representations of the data. Finally, a DNN with and without the encoded representations was built to detect severity, with the neural network without the encoded representation maintaining the best performance metrics. This DNN can be used to predict the severity of an accident. Future applications of this model could be in city planning and road construction. Using models to predict the severity of future accidents based on geographic and POI city planners could predict the risk of severe accidents using their road plans.

\bibliographystyle{splncs04}
\bibliography{example_paper}

\begin{thebibliography}{10}
\providecommand{\url}[1]{\texttt{#1}}
\providecommand{\urlprefix}{URL }
\providecommand{\doi}[1]{https://doi.org/#1}

\bibitem{CramersV}
Akoglu, H.: User's guide to correlation coefficients. Turkish Journal of
  Emergency Medicine  \textbf{18} (2018),
  \url{https://www.sciencedirect.com/science/article/pii/S2452247318302164}

\bibitem{bahadur2019dimension}
Bahadur, N., Paffenroth, R.: Dimension estimation using autoencoders (2019)

\bibitem{9225336}
Behura, A., Behura, A.: Road accident prediction and feature analysis by using
  deep learning. In: 2020 11th International Conference on Computing,
  Communication and Networking Technologies (ICCCNT). pp.~1--7. IEEE (2020)

\bibitem{castillo2019legislation}
Castillo-Manzano, J.I., Castro-Nu{\~n}o, M., L{\'o}pez-Valpuesta, L., Pedregal,
  D.J.: From legislation to compliance: The power of traffic law enforcement
  for the case study of spain. Transport policy  \textbf{75}, ~1--9 (2019)

\bibitem{8530861}
Chen, C., Fan, X., Zheng, C., Xiao, L., Cheng, M., Wang, C.: Sdcae: Stack
  denoising convolutional autoencoder model for accident risk prediction via
  traffic big data. In: 2018 Sixth Intl. Conf. on Advanced Cloud and Big Data.
  pp. 328--333. IEEE (2018)

\bibitem{chen2019global}
Chen, S., Kuhn, M., Prettner, K., Bloom, D.E.: The global macroeconomic burden
  of road injuries: estimates and projections for 166 countries. The Lancet
  Planetary Health  \textbf{3}(9),  e390--e398 (2019)

\bibitem{favretto2018driving}
Favretto, D., Visentin, S., Stocchero, G., Vogliardi, S., et~al.: Driving under
  the influence of drugs: Prevalence in road traffic accidents in italy and
  considerations on per se limits legislation. Traffic injury prevention
  \textbf{19}(8),  786--793 (2018)

\bibitem{gaber2017traffic}
Gaber, M., Wahaballa, A.M., Othman, A.M., Diab, A.: Traffic accidents
  prediction model using fuzzy logic: Aswan desert road case study. J. Eng.
  Sci. Assiut Univ  \textbf{45}, ~2844 (2017)

\bibitem{10.1145/3357384.3357829}
Huang, C., Zhang, C., Dai, P., Bo, L.: Deep dynamic fusion network for traffic
  accident forecasting. In: Proceedings of the 28th ACM International
  Conference on Information and Knowledge Management. pp. 2673--2681 (2019)

\bibitem{ihueze2018road}
Ihueze, C.C., Onwurah, U.O.: Road traffic accidents prediction modelling: An
  analysis of anambra state, nigeria. Accident Analysis \& Prevention
  \textbf{112},  21--29 (2018)

\bibitem{8687542}
Liyong, W., Vateekul, P.: Improve traffic prediction using accident embedding
  on ensemble deep neural networks. In: 2019 11th International Conference on
  Knowledge and Smart Technology (KST). pp. 11--16. IEEE (2019)

\bibitem{5203263}
Lv, Y., Tang, S., Zhao, H.: Real-time highway traffic accident prediction based
  on the k-nearest neighbor method. In: 2009 international conference on
  measuring technology and mechatronics automation. vol.~3, pp. 547--550. IEEE
  (2009)

\bibitem{5620615}
Lv, Y., Haixia, Z., Xing-lin, Z., Ming, L., Jie, L.: Research on accident
  prediction of intersection and identification method of prominent accident
  form based on back propagation neural network. In: 2010 International
  Conference on Computer Application and System Modeling (ICCASM 2010). vol.~1,
  pp. V1--434. IEEE (2010)

\bibitem{mcinnes2020umap}
McInnes, L., Healy, J., Melville, J.: Umap: Uniform manifold approximation and
  projection for dimension reduction (2020)

\bibitem{RoadTrafficAccidentSeveritySouthAfrica}
Mokoatle, M., Vukosi~Marivate, D., Michael Esiefarienrhe~Bukohwo, P.:
  Predicting road traffic accident severity using accident report data in south
  africa. In: Proc. of the 20th Annual Intl. Conf. on Digital Government
  Research. pp. 11--17 (2019)

\bibitem{CornellCitation1}
Moosavi, S., Samavatian, M.H., Parthasarathy, S., Ramnath, R.: A countrywide
  traffic accident dataset. arXiv preprint arXiv:1906.05409  (2019)

\bibitem{CornellCitation2}
Moosavi, S., Samavatian, M.H., Parthasarathy, S., Teodorescu, R., Ramnath, R.:
  Accident risk prediction based on heterogeneous sparse data: New dataset and
  insights. In: Proceedings of the 27th ACM SIGSPATIAL International Conference
  on Advances in Geographic Information Systems. pp. 33--42 (2019)

\bibitem{ren2018deep}
Ren, H., Song, Y., Wang, J., Hu, Y., Lei, J.: A deep learning approach to the
  citywide traffic accident risk prediction. In: 2018 21st International
  Conference on Intelligent Transportation Systems (ITSC). pp. 3346--3351. IEEE
  (2018)

\bibitem{rivas2019deep}
Rivas, P., Rivas, E., Velarde, O., Gonzalez, S.: Deep sparse autoencoders for
  american sign language recognition using depth images. In: Proceedings on the
  International Conference on Artificial Intelligence (ICAI). pp. 438--444
  (2019)

\bibitem{sameen2017severity}
Sameen, M.I., Pradhan, B.: Severity prediction of traffic accidents with
  recurrent neural networks. Applied Sciences  \textbf{7}(6), ~476 (2017)

\bibitem{sheng2018effect}
Sheng, R., Zhong, S., Barnett, A.G., Weiner, B.J., Xu, J., Li, H., Xu, G., He,
  T., Huang, C.: Effect of traffic legislation on road traffic deaths in
  ningbo, china. Annals of epidemiology  \textbf{28}(8),  576--581 (2018)

\bibitem{8367975}
{Singh}, D., {Mohan}, C.K.: Deep spatio-temporal representation for detection
  of road accidents using stacked autoencoder. IEEE Transactions on Intelligent
  Transportation Systems  \textbf{20}(3),  879--887 (2019).
  \doi{10.1109/TITS.2018.2835308}

\bibitem{10.1145/3356995.3364535}
Takimoto, Y., Tanaka, Y., Kurashima, T., Yamamoto, S., Okawa, M., Toda, H.:
  Predicting traffic accidents with event recorder data. In: Proc. of the 3rd
  ACM SIGSPATIAL Intl. Workshop on Prediction of Human Mobility. pp. 11--14
  (2019)

\bibitem{BigDataTraffic}
Tantaoui, M., Laanaoui, M.D., Kabil, M.: Real-time prediction of accident using
  big data system. In: Proceedings of the 3rd International Conference on
  Networking, Information Systems \& Security. pp.~1--5 (2020)

\bibitem{8056908}
Wenqi, L., Dongyu, L., Menghua, Y.: A model of traffic accident prediction
  based on convolutional neural network. In: 2017 2nd IEEE International
  Conference on Intelligent Transportation Engineering (ICITE). pp. 198--202.
  IEEE (2017)

\bibitem{8842686}
Xia, X.L., Nan, B., Xu, C.: Real-time traffic accident severity prediction
  using data mining technologies. In: 2017 International Conference on Network
  and Information Systems for Computers (ICNISC). pp. 242--245. IEEE (2017)

\end{thebibliography}

\end{document}